\documentclass{article}

\usepackage{microtype}
\usepackage{graphicx}
\usepackage{caption}
\usepackage{subcaption}
\usepackage{booktabs} 
\usepackage{multirow}
\usepackage{xspace}
\usepackage{bbding}
\usepackage{amsfonts}
\usepackage{makecell}
\usepackage{enumitem}
\usepackage{amsmath}

\usepackage{hyperref}

\newcommand{\NAME}{$\mathtt{ShiftNAS}$\xspace}

\usepackage[accepted]{icml2021}

\icmltitlerunning{ShiftNAS: Towards Automatic Generation of Advanced Mulitplication-Less Neural Networks}

\begin{document}

\twocolumn[
\icmltitle{ShiftNAS: Towards Automatic Generation of \\ Advanced Mulitplication-Less Neural Networks}



\icmlsetsymbol{equal}{*}

\begin{icmlauthorlist}
\icmlauthor{Xiaoxuan Lou}{one}
\icmlauthor{Guowen Xu}{one}
\icmlauthor{Kangjie Chen}{one}
\icmlauthor{Guanlin Li}{one}
\icmlauthor{Jiwei Li}{two}
\icmlauthor{Tianwei Zhang}{one}
\end{icmlauthorlist}

\icmlaffiliation{one}{Nanyang Technological University}
\icmlaffiliation{two}{Shannon.AI}


\icmlkeywords{Machine Learning, ICML}

\vskip 0.3in
]



\printAffiliationsAndNotice{} 

\begin{abstract}
 Multiplication-less neural networks significantly reduce the time and energy cost on the hardware platform, as the compute-intensive multiplications are replaced with lightweight bit-shift operations. However, existing bit-shift networks are all directly transferred from state-of-the-art convolutional neural networks (CNNs), which lead to non-negligible accuracy drop or even failure of model convergence. 
 To combat this, we  propose \NAME, the first framework tailoring Neural Architecture Search (NAS) to substantially reduce the accuracy gap between bit-shift neural networks and their real-valued counterparts. Specifically, we pioneer dragging NAS into a \textit{shift-oriented search space} and endow it with the robust \textit{topology-related search strategy} and custom \textit{regularization and stabilization}. As a result, our \NAME breaks through the incompatibility of traditional NAS methods for bit-shift neural networks and achieves more desirable performance in terms of accuracy and convergence. Extensive experiments demonstrate that \NAME sets a new state-of-the-art for bit-shift neural networks, where the accuracy increases (1.69$\sim$8.07)\% on CIFAR10, (5.71$\sim$18.09)\% on CIFAR100 and (4.36$\sim$67.07)\% on ImageNet, especially when many conventional CNNs fail to converge on ImageNet with bit-shift weights.

\end{abstract}

\section{Introduction} \label{sec_intro}
In recent years, large-scale commercial applications based on convolutional neural networks (CNNs) have prompted researchers to design more efficient networks, which can be deployed on platforms with limited resource budget, such as mobile or IoT devices. Early works utilized network quantization \cite{cheng2017survey}  to achieve this goal, which replaces high-precision 32-bit floating-point model parameters with lower-precision smaller bit-width representations. It can reduce the computational cost of model execution, but also suffer from a non-negligible performance degradation, especially on complex datasets (e.g., ImageNet). To address this issue, recent works \cite{zhou2017incremental,elhoushi2021deepshift} turned to use binary bit shifts rather than simple quantized bits to replace floating-point model parameters.


The key insight of these solutions is that multiplying an element by a power of 2 is mathematically equivalent to a bit shfit operation on it, which is computationally much cheaper and hardware-friendly. Based on this, researchers designed different types of \textit{bit-shift techniques}, e.g., INQ \citep{zhou2017incremental} and DeepShift \cite{elhoushi2021deepshift}, which replace the multiplications in neural networks with bit shift operations. These solutions show promising overhead reduction in model execution. 
However, existing bit-shift networks are all directly transferred from conventional CNNs, e.g., ResNets \cite{he2016deep} and VGG \cite{simonyan2014very}. Since these conventional CNNs are all designed for the continuous real-valued domain, such direct conversion can restrict the potential of bit-shift techniques, causing less optimal network architecture with non-trivial accuracy drop. 



To overcome this limitation, we aim to automatically generate the optimal bit-shift network architectures with the best performance. This is inspired by the Neural Architecture Search (NAS) technique, which can automatically identify the satisfactory network architecture for a given task. The searched models have shown better performance than carefully hand-crafted models \cite{liu2018darts, chen2019progressive}. One straightforward way is to apply the conventional NAS methods to obtain a model in the real-valued domain, and then transfer it to the bit-shift network. However, similar as the manually-crafted networks, such strategy also leads to the sub-optimal results due to the semantic gap between real and bit-shift domains (Sections~\ref{sec_overview} and \ref{sec_abst}). 


For the first time, we present \NAME, a novel methodology to automatically search for the optimal bit-shift network architectures directly, aiming to reduce the accuracy drop from the state-of-the-art real-valued models. 
Moreover, the introduction of bit-shift operations can significantly reduce the searching, training and inference cost, which can facilitate the deployment of large models on dedicated hardware. Specifically, \NAME contains 3 novel components. (1) \textit{Shift-oriented search space}. While existing NAS techniques mainly focus on the real-valued domain, we are the first to construct a new search space composed of bit-shift operations and design the corresponding forward and backward pass. (2) \textit{Topology-related search strategy}. Since bit-shift networks tend to have faster gradient descent \cite{elhoushi2021deepshift}, they are more vulnerable with the conventional gradient-based NAS techniques, i.e., searched networks can be dominated by skip connections \cite{liu2018progressive}. Therefore, we design a robust search strategy based on DOTS \cite{gu2021dots} to mitigate this issue, which adopts group operation search and also considers topology search. (3) \textit{Search regularization and stabilization}. To further improve the performance of searched network architectures, we adopt two approaches to regularize and stabilize the search procedure, including the modified L2 regularization for shift parameters and learning rate reset scheme.

The networks searched by \NAME show much better performance than conventional CNNs in the bit-shift domain, especially when many CNNs fail to converge on large datasets (e.g., ImageNet) with bit-shift weights. \NAME achieves an accuracy improvement of (1.69$\sim$8.07)\% on CIFAR10, (5.71$\sim$18.09)\% on CIFAR100 and (4.36$\sim$67.07)\% on ImageNet, with more compact parameter sizes and smaller numbers of operation computations. 
Compared with previous NAS methods, networks from \NAME are more compatible to the bit-shift domain, which lead to smaller accuracy drop from the complex real-valued models. More importantly, \NAME consumes less computing resources and time as it directly searches with the bit-shift weights.

\section{Preliminaries} \label{sec_back}

\subsection{Bit-shift Network Quantization} \label{sec_bitshift}
Conventional CNN quantization techniques \cite{rastegari2016xnor} quantize the 32-bit floating-point model weights to a smaller number of bits. Recently, bit shift approaches are introduced \citep{zhou2017incremental, elhoushi2021deepshift}, which round model weights to the powers of 2 so that the intensive multiplications on weights can be achieved with cheaper binary bit shifts. Formally, given a number $x$ and a rounded model weight $2^p$, their multiplication is mathematically equivalent to shifting $p$ bits of $x$:
\begin{equation}
2^{p}x = \left\{
\begin{array}{ll}
x <<  p & {if} ~~ p > 0 \\
x >>  p  & {if} ~~ p < 0 \\
x    & {if} ~~ p = 0
\end{array}
\right .
\end{equation}
Since model weights can be either positive or negative for input feature extraction, while $2^{p}$ is always positive, \emph{sign flip} is thus introduced to represent the signs of weight values. 
This operation is achieved with a ternary sign operator $s \in \{-1,0,+1\}$:
\begin{equation}
flip(x,s) = \left\{
\begin{array}{ll}
-x & {if} ~~  s = -1 \\
0   & {if} ~~  s = 0 \\
x    & {if} ~~  s = +1
\end{array}
\right .
\end{equation}
Hence, we can replace the weight matrix $W$ in the model as: $W = flip(2^{P},S)$, where $P$ is the shift matrix and $S$ is the sign matrix. Both bit shift and sign flip are computationally cheap, as the former is the fundamental operation in modern processors and the latter just computes 2's complement of a number. Therefore, such weight replacement can efficiently reduce the computation cost of CNN model execution.

\subsection{Neural Architecture Search}
NAS has gained great popularity in recent years, due to its capability of building machine learning pipelines with high efficiency and automation. Early methods used reinforcement learning \cite{zoph2016neural} and evolutionary algorithms \cite{real2019regularized} to search for optimal network architectures for a given task, which normally takes thousands of GPU hours. Recent works tended to use gradient-based strategy \cite{liu2018darts} that can reduce the search cost to a few hours. Most gradient-based methods aim at searching for optimal cell structures, since stacking cells as a model is more efficient than searching the whole network architecture. 
Formally, a cell is represented as a directed cyclic graph (i.e., \emph{supernet}) with $N$ nodes $\{x_i\}_{i=1}^N$, including two inputs and one output, and several intermediate nodes. The $j$-th intermediate node $x_j$ connects to all previous nodes $x_i$ through the edge $(i,j)$. The operation choice over the edge $(i,j)$ can be relaxed as:
\begin{equation}\label{eq_oprelax}
    \begin{aligned}
         & {\overline o}^{(i,j)}(x) = \sum \limits_{o \in \mathcal{O}} \alpha_o^{(i,j)} o^{(i,j)}(x_i),   \\
         & \alpha_o^{(i,j)} = \frac{exp({\alpha'}_o^{(i,j)})}{\sum \limits_{o' \in \mathcal{O}} exp({\alpha'}_{o'}^{(i,j)})}    
    \end{aligned}
\end{equation}
where $o \in \mathcal{O}$ and $\mathcal{O}$ denotes the search space of candidate operations. $\alpha_o^{(i,j)}$ is the trainable weight for each operation on the edge $(i,j)$, which is normalized with the softmax function. Therefore, the feature map of node $x_j$ can be computed by adding all results from its predecessors $x_i$: 
\begin{equation}
    x_j = \sum_{i<j} \overline o^{(i,j)} (x_i)
\end{equation}
Let $\mathcal{L}_{train}$ and $\mathcal{L}_{val}$ denote the model loss on the training and validation sets.  A bi-level optimization is applied to the operation weight $\alpha$ and network weight $w$ as:
\begin{equation}
\begin{aligned}
    & \mathop{min}\limits_{\alpha}~~\mathcal{L}_{val}(w^*(\alpha), \alpha), \\
    & s.t.~~w^*(\alpha) = arg\mathop{min}\limits_{w}(\mathcal{L}_{train}(w,\alpha))
\end{aligned}
\end{equation}
The final model architecture can be derived from the trained operation weight $\alpha$ by retaining operations with the largest weight and pruning edges with the smaller weight.

\begin{figure*}[t]
    \centering
    \includegraphics[width=0.8\linewidth]{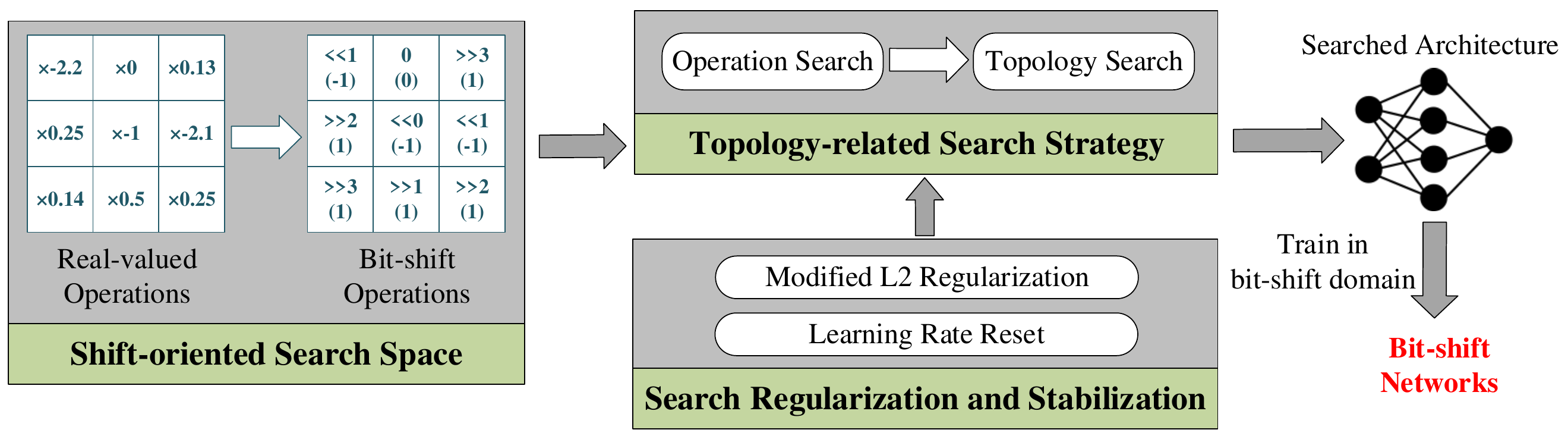}
    \vspace{-10pt}
    \caption{The overview of ShiftNAS}
    \label{fig_overview}
    \vspace{-10pt}
\end{figure*}

\section{Overview of \NAME} \label{sec_overview}
The main idea of \NAME is to automatically generate well-performed bit-shift networks with high efficiency.
It has multiple challenges to apply exiting NAS techniques for searching bit-shift networks: 

\textbf{Design of shift-oriented search space.} Given that existing NAS methods mainly focus on the real-valued models, their search spaces are also designed for real domain, which is not applicable to bit-shift models. Specifically, a conventional NAS search space normally consists of multiple manually defined operations, such as \emph{dilated convolutions} and \emph{separable convolutions}. To build the shift-oriented search space, we need to transfer these operations from the real domain into the bit-shift domain, in which the forward pass and backward pass need to be carefully designed. 

\textbf{Dominance of skip connections.} While most of recent NAS methods adopt the gradient-based search strategy (i.e., DARTS \cite{liu2018darts}), it has a big drawback: the searched networks are easy to be dominated by skip connections \cite{chen2019progressive}, as the strategy prefers the fastest way of gradient descent. Unfortunately, searching in the bit-shift domain inherits and amplifies this drawback, which would lead to the "cell collapsing" of searched architectures. Hence, a new search strategy considering both the model operations and topology should be adopted.

\noindent\textbf{Less robust search procedure.} Replacing floating-point weights with bit shifts brings fast computations, but also results in the accuracy drop and difficulty of model training. Specifically, the introduced shift parameters and sign flips should be well regularized to avoid errors in the gradient descent. Besides, since bit-shift operations are extremely sensitive to a large learning rate, the selection and scheduling of the learning rate should also be carefully crafted. 

We design a novel NAS technique \NAME to address the above challenges. Figure \ref{fig_overview} shows the overview of our methodology, which consists of three key components: 
\begin{itemize}[leftmargin=*, itemsep=0pt, topsep=0pt]
    \item \emph{Shift-oriented search space.} This new search space consists of 8 operations, which are converted from the real domain to bit-shift domain. 
    \item \emph{Topology-related search strategy.} This new strategy considers the optimal combination of model operations and topology, which can efficiently mitigate the dominance of skip connections.
    \item \emph{Search regularization and stabilisation.} Two approaches are proposed to regularize and stabilize the search procedure: applying a modified L2 regularization for shift parameters and resetting the learning rate during search. 
\end{itemize}

\section{Methodology} \label{sec_method}
In this section, we describe the detailed mechanism of each component in \NAME. 

\subsection{Shift-oriented Search Space}
\label{sec:search-space}
Following previous NAS works (e.g., DARTS \cite{liu2018darts}), we adopt 8 operations as our operation search space $\mathcal{O}$: $3 \times 3$ and $5 \times 5$ dilated convolutions, $3 \times 3$ and $5 \times 5$ separable convolutions, $3 \times 3$ max pooling, $3 \times 3$ average pooling, identity (skip) and the zero\footnote{zero means no connection between two nodes.}. To construct a shift-oriented search space, we propose a bit-wise shift technique following DeepShift \cite{elhoushi2021deepshift} and transfer these operations into the bit-shift domain.

\noindent\textbf{Grouping candidate operations.} 
Since not every candidate operation needs to be transferred into the bit-shift version, e.g., the identity and pooling, we first divide 8 candidate operations (excluding \emph{zero}) into two groups. The first group $\mathcal{O}_{c}$ contains four \emph{convolution} operations, which involve dense multiplications. The second group $\mathcal{O}_{t}$ contains the remaining operations, which mainly focus on the model topology, such as \emph{skip} and \emph{pooling}. The entire search space is denoted as 
$\mathcal{O} = \{\mathcal{O}_{c}, \mathcal{O}_{t}\}$. 
To construct the shift-oriented search space, we just need to transfer operations in $\mathcal{O}_{c}$ into the bit-shift domain, and keep operations in $\mathcal{O}_{t}$ unchanged. Note that this operation group scheme will also be adopted in the topology-related search strategy (Section \ref{sec_topo}).

\noindent\textbf{Replacement of operation weights.} 
As introduced in Section \ref{sec_bitshift}, quantization of bit-shift networks can be implemented by replacing the floating-point model weights with two parameters: bit shift $P$ and sign flip $S$. Hence, the weights $w$ of operations in $\mathcal{O}_{c}$ need to be replaced with the trainable parameters $(P,S)$, which is formulated as below: 
\begin{equation}
\begin{aligned}
& \overline P = round(P) \\
& \overline S = sign(round(S)) \\
& w = flip(2^{\overline P}, \overline S)
\end{aligned}
\end{equation}
where $\overline P$ is the rounded shift matrix and $\overline S$ is the rounded sign matrix. Note that the function \emph{sign} generates a ternary value, and can be represented as:
\begin{equation}
sign(s) = \left\{
\begin{array}{ll}
-1 & {if} ~~ s \leq -0.5 \\
0   & {if} ~~ -0.5 < s < 0.5 \\
+1    & {if} ~~ s \geq 0.5
\end{array}
\right .
\end{equation}

\noindent\textbf{Designing forward and backward pass.} 
Different from some previous works \cite{zhou2017incremental} which just rounded the trained models into the bit-shift domain, our goal is to directly search and train the model in the shift domain. So we need to design and implement the forward and backward pass of shift operations. 
With the transferred weights $w = flip(2^{\overline P}, \overline S)$, the forward pass for convolutions in $\mathcal{O}_{c}$ can be represented as: $Y = w*X = flip(2^{\overline P}, \overline S)*X + b$, where $(X,Y)$ denote the operation input and output, and $b$ denotes the bias. The gradients of the backward pass can be formulated as: 
\begin{equation}
\begin{aligned}
& \frac{\partial \mathcal{L}}{\partial X} = \frac{\partial \mathcal{L}}{\partial Y} \frac{\partial Y}{\partial X} = \frac{\partial \mathcal{L}}{\partial Y} w^T  \\
& \frac{\partial \mathcal{L}}{\partial P} = \frac{\partial \mathcal{L}}{\partial Y} \frac{\partial Y}{\partial w} \frac{\partial w}{\partial \overline P} \frac{\partial \overline P}{\partial P} \\
& \frac{\partial \mathcal{L}}{\partial S} = \frac{\partial \mathcal{L}}{\partial Y} \frac{\partial Y}{\partial w} \frac{\partial w}{\partial \overline S} \frac{\partial \overline S}{\partial S} \\
& \frac{\partial \mathcal{L}}{\partial b} = \frac{\partial \mathcal{L}}{\partial Y}
\end{aligned}
\end{equation}
where $\mathcal{L}$ denotes the model loss. 

We use the straight through estimators \cite{yin2019understanding} to compute the derivatives of the round and sign functions as: $\frac{\partial {round}(x)}{\partial x} \approx 1$ and $\frac{\partial sign(x)}{\partial x} \approx 1$. For the sign flip function, we have: $\frac{\partial {flip}(x,s)}{\partial x} \approx flip(x,s)$ and $\frac{\partial flip(x,s)}{\partial s} \approx 1$. With these estimations, we can set $\frac{\partial \overline P}{\partial P} \approx 1$ and $\frac{\partial \overline S}{\partial S} \approx 1$, and then obtain the following expressions:
\begin{equation}
\begin{aligned}
    & \frac{\partial w}{\partial \overline S} = \frac{\partial flip(2^{\overline P}, \overline S)}{\partial \overline S} \approx 1 \\
    & \frac{\partial w}{\partial \overline P}  = \frac{\partial flip(2^{\overline P}, \overline S)}{\partial \overline P} = \frac{\partial flip(2^{\overline P}, \overline S)}{\partial 2^{\overline P}} \frac{\partial 2^{\overline P}}{\partial \overline P} \\
    & ~~~~~~\approx flip(2^{\overline P}, \overline S) 2^{\overline P} ln2 = w  2^{\overline P}  ln2
\end{aligned}
\end{equation}
As a result, the gradients of the trainable parameters $(P,S)$ with respect to the model loss $\mathcal{L}$ are set to:
\begin{equation}
\begin{aligned}
    & \frac{\partial \mathcal{L}}{\partial P} \approx \frac{\partial \mathcal{L}}{\partial Y} \frac{\partial Y}{\partial w} w  2^{\overline P}  ln2 \\
    & \frac{\partial \mathcal{L}}{\partial S} \approx \frac{\partial \mathcal{L}}{\partial Y} \frac{\partial Y}{\partial w} 
\end{aligned}
\end{equation}
Based on the above constructed forward and backward pass of bit-shift operations, we can achieve searching and training a NAS model in the bit-shift domain. 

\subsection{Topology-related Search Strategy} \label{sec_topo}
The dominance of skip connections caused by the gradient-based search strategy is a major restriction for applying NAS techniques to quantized networks \cite{bulat2020bats}. Besides, ignoring the model topology during search in some NAS methods also limits the generation of optimal network architectures. 
Hence, we propose an advanced search strategy based on DOTS \cite{gu2021dots}, which considers both the operation search and topology search. This strategy can efficiently suppress the dominance of skip-connections and also improve the performance of searched networks. 

\noindent\textbf{Operation search.}
As introduced in Section \ref{sec:search-space}, the 8 candidate operations in the shift-oriented search space can be divided into two groups: $\mathcal{O}_{t}$ contains topology-related operations that can explicitly affect the model topology (e.g., \emph{skip}), while operations in $\mathcal{O}_{c}$ do not have such impact. Therefore, 
the operation search space $\mathcal{O}$ is split into two subspaces $\mathcal{O} = \{\mathcal{O}_{t}, \mathcal{O}_{c}\}$, and each operation subspace is relaxed to be continuous independently as shown in Eq.(\ref{eq_oprelax}). Then a bi-level optimization is applied to train the model weight $w$ and operation weight $\alpha$. 
With the trained $\alpha$, we retain the operation with the maximum weight in each operation subspace, which can be formulated as: 
\begin{equation}
\begin{aligned}
     & o_{t}^{(i,j)} = arg\mathop{max}\limits_{o_{t} \in \mathcal{O}_{t}} \alpha_{o_{t}}^{(i,j)} \\
     & o_{c}^{(i,j)} = arg\mathop{max}\limits_{o_{c} \in \mathcal{O}_{c}} \alpha_{o_{c}}^{(i,j)} \\
\end{aligned}
\end{equation}
Such group operation scheme can avoid the elimination of potential topology choices during the operation search, which then allows the subsequent topology search to find out the optimal topology. 
Finally, all the retained operations are collected to construct a new operation search space $\mathcal{O}_N=\{o_{t}^{(i,j)}, o_{c}^{(i,j)}\}$ on each edge $(i,j)$, which is used for the topology search. 

\noindent\textbf{Topology search.}
The previous operation search step aims to determine the best operations on each edge. In this topology search step, we try to search for the optimal combinations of model edges. It can well prevent \emph{skips} from dominating the searched model topology. 

First, a topology search space is constructed. Following previous works, we restrict two input edges for each node in the cell \emph{supernet}, so the topology search space $\mathcal{E}_{x_j}$ for node $x_j$ can be represented as a set of all possible pairwise combinations of its incoming edges: $\mathcal{E}_{x_j} = \{ \langle (i_1,j),(i_2,j) \rangle | 0<i_1<i_2<j \}$. The topology search space contains $C_n^2=\frac{n!}{2!(n-2)!}$ candidates, where $n$ denotes the number of incoming edges for node $x_j$. Similar with the operation search, we also relax the topology search space $\mathcal{E}_{x_j}$ to be continuous:
\begin{equation}\label{eq_topo}
    \beta_{x_j}^c = \frac{exp({\beta'}_{x_j}^c/T_{\beta})}{\sum \limits_{c' \in \mathcal{E}_{x_j}} exp({\beta'}_{x_j}^{c'}/T_{\beta}))}
\end{equation}
where $\beta_{x_j}^c$ is the topology weight that denotes the normalized probability of the edge combination $c \in \mathcal{E}_{x_j}$. $T_{\beta}(t)=T_0 \theta^t$ is the temperature for architecture annealing, which can efficiently bridge the optimization gap between the supernet and child networks \cite{xie2018snas}. 

Then, the importance weight $\gamma^{(i,j)}$ for each edge $(i,j)$ can be computed from those combinations containing this edge, which can be formulated as:
\begin{equation}
    \gamma^{(i,j)} = \sum \limits_{c \in \mathcal{E}_{x_j}, (i,j) \in c} \frac{1}{N(c)}\beta_{x_j}^c
\end{equation}
where $N(c)$ is the number of edges in the edge combination $c$. As a result, the feature map of node $x_j$ can be obtained by summing all the incoming edges weighted by the edge importance weight $\gamma^{(i,j)}$:
\begin{equation}
    x_j=\sum \limits_{i<j} \gamma^{(i,j)} {\overline o}^{(i,j)}(x_i)
\end{equation}
where ${\overline o}^{(i,j)}(x_i)$ denotes the mixed operations on edge $(i,j)$ obtained from the operation search. In the topology search, as the number of candidate operations is largely reduced (i.e., 2 in $\mathcal{O}_N$), we can directly use the one-level optimization to update three weights $(w, \alpha, \beta)$ in the search.

\noindent\textbf{Determining the architecture.} After the operation and topology search, we select the edge combination $c$ with the maximal weight in topology weight $\beta$ to construct the model topology, and then attach to each edge the operation with the maximal weight in the operation weight $\alpha$. 

\subsection{Search Regularization and Stabilisation}\label{sec_rest}
Based on the shift-oriented search space and topology-related search strategy, an efficient bit-shift network architecture can be identified for each specific task automatically. However, the adoption of bit-shift weights makes the architecture search much more unstable and also leads to more difficult model training. The search process usually converges to a sub-optimal solution, sometimes even cannot converge. We propose two approaches to regularize and stabilize the optimization of the three trainable weights in the search procedure: network weight $w$, operation weight $\alpha$ and topology weight $\beta$. 

\begin{figure}[t]
    \centering
    \includegraphics[width=0.9\linewidth]{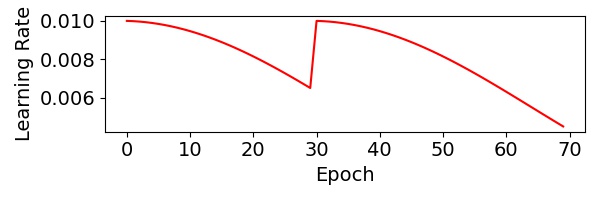}
    \vspace{-15pt}
    \caption{Learning rate curve in the architecture search}
    \label{fig_lr}
    \vspace{-10pt}
\end{figure}

For the optimization of the network weight $w$, note that $w$ consists of the bitwise shift $P$ and sign flip $S$, i.e., $w \leftarrow \{P,S\}$. We use a \emph{modified L2 regularization} term to regularize the gradient descent of $P$, which is defined as $\sum W^2 = \sum (2^P S)^2$ rather than the conventional formulation $\sum (P^2 + S^2)$. While most weights in a trained model are rarely larger than 1 (i.e., $|2^P| < 1$), the range of the value of $P$ is also empirically set to be smaller than 0. As a negative parameter, a smaller $P$ instead leads to a larger $P^2$, which gives a reverse activation to the training loss. Hence, the regularization term should be modified to avoid misguiding the direction of the gradient descent. Formally, the regularized loss $\mathcal{L}'$ can be formulated as:
\begin{equation}
    \mathcal{L}' = \mathcal{L} + \frac{\lambda}{2} \sum (2^P S)^2
\end{equation}
where $\mathcal{L}$ denotes the original model loss and $\lambda$ is the fixed weight decay. Our experiments in Section \ref{sec_abst} show that this modified \emph{L2 regularization} improves the accuracy of searched architectures. 

To stabilize the optimization of the operation weight $\alpha$ and topology weight $\beta$, in addition to using the temperature regularization in Eq.(\ref{eq_topo}), we also carefully implement a \emph{learning rate reset} scheme. Since bit-shift networks are extremely sensitive to large learning rates, we need to use a much smaller initial learning rate than that in previous NAS techniques 
to avoid model convergence failure. Besides, while the topology search in DOTS adopts the annealed learning rate from the previous operation search step, we find that resetting the learning rate to an initial value again at the start of topology search allows to obtain a better network architecture. Figure \ref{fig_lr} shows the learning rate curve in the search with the cosine annealing: the learning rate is reset at the 30\textit{th} epoch, when the topology search starts.

\begin{table*}[t]
\centering
\resizebox{\linewidth}{!}{
\begin{tabular}{lcccccc}
    \hline
    \hline
    \multirow{2}{*}{\textbf{Architecture}} & \textbf{Top-1 Acc. (\%)} & \textbf{Params (M)} & \textbf{Top-1 Acc. (\%)} & \textbf{Params (M)} & \textbf{Search Cost}  & \textbf{Search}\\
        &  \textbf{CIFAR10}  &   \textbf{CIFAR10}  &  \textbf{CIFAR100}   &  \textbf{CIFAR100} & \textbf{(GPU-days)} & \textbf{Method} \\
    \hline
    ResNet18 \cite{he2016deep} &  93.20   &  11.2  &  69.11  &  11.2  & -   & -   \\
    ResNet20 \cite{he2016deep} &  88.84   &  0.3  &  60.12   &  0.3   & -   & -  \\
    ResNet50 \cite{he2016deep} & 93.89 &  23.6  &   70.64  &  23.6   & -   & -  \\
    ResNet56 \cite{he2016deep} &  91.11   & 0.9  &   65.57  & 0.9    & -  & -   \\
    ResNet101 \cite{he2016deep} &  93.43  & 42.8   &  69.18   &   42.8    & -  & - \\
    ResNet152 \cite{he2016deep} & 93.17    & 58.5   &  65.58  &   58.5     & - & - \\
    MobileNet-v2 \cite{sandler2018mobilenetv2} & 92.64  & 2.4  &    70.24       &    2.4  & -  & -  \\
    VGG19 \cite{simonyan2014very} & 91.57  &  20.1  &   64.88   &  20.1   & -   & -  \\
    ShuffleNet-v2 \cite{ma2018shufflenet} &   87.51   &  0.5 &  58.26    &   0.5    & -  & - \\
    \hline
    NASNet \cite{zoph2018learning} & 95.28  &  3.3  &  75.33   & 3.3  & 1800  & RL    \\
    AmoebaNet \cite{real2019regularized} &  95.22  &  2.3  &  75.05  &  2.3  &   3150  & EA \\
    DARTS-v1 \cite{liu2018darts} &       94.39     &  3.2  &   74.93        &  3.2 &   0.4  & GD  \\
    DARTS-v2 \cite{liu2018darts} &   94.80   &  3.5  &  75.17   &   3.5 &    0.4 & GD \\
    GDAS \cite{dong2019searching} &   94.62  &  2.5  &   74.87   &   3.4  &   0.2 & GD \\
    P-DARTS \cite{chen2019progressive} &   94.21  &  3.4  &  74.54   & 3.6  &   0.3  & GD  \\
    DARTS- \cite{xu2019pc} &        93.87    &  3.4  &   70.85        & 3.5  &   0.4  & GD  \\
    DOTS \cite{gu2021dots} &   95.12   &  3.7  &  75.05  &   4.2   &  0.3 & GD \\
    \hline
    \NAME (Best) &  \textbf{95.58}  & 3.3  &   \textbf{76.35}   & 3.8 & 0.23* & GD \\
    \NAME (Avg)$\dagger$ &  95.43$\pm$0.12  & 3.3  &   76.08$\pm$0.23   & 3.8 & 0.23* & GD \\
    \hline
    \hline
\end{tabular}}
\caption{Evaluation results on CIFAR10/100. The results of conventional CNNs are obtained by running open code of DeepShift \cite{elhoushi2021deepshift}. $\dagger$: The results are computed from four individual runs with random seeds. *: The search cost can be much smaller on the dedicated hardware as we emulate the bit-shift operations with software.}\label{tab_cifar}
\vspace{-10pt}
\end{table*}

\section{Evaluation} \label{sec_eval}
We implement \NAME with Pytorch. Following previous works \cite{elhoushi2021deepshift,zhou2017incremental}, we emulate the precision of an actual bit-shift hardware implementation by rounding the operation input and bias to the 32-bit fixed-point format precision (16-bit for the integer part and 16-bit for the fraction part). The shift parameter $P$ is constrained in [-15, 0], i.e., the absolute value of the model weight is within $[2^{-15},1]$, which only needs 4 bits to represent. The model weight also needs an extra bit to denote its sign $S$. 

We run evaluations on CIFAR10, CIFAR100 and ImageNet datasets. We comprehensively compare \NAME with a variety of state-of-the-art CNN models (e.g., ResNet, VGG, MobileNet, ShuffleNet, GoogleNet, SqueezeNet) and NAS models (e.g., NASNet, AmoebaNet, DARTS, GDAS, DOTS). For fair comparisons, these baseline models are trained in the bit-shift domain, if not mentioned otherwise.


\subsection{Evaluation on CIFAR}
\noindent\textbf{Search settings.} 
The entire search process on CIFAR 10/100 consists of two steps: operation search for 30 epochs and then topology search for 40 epochs. The network skeleton consists of 8 cells (6 normal cells and 2 reduction cells) with the initial channel size of 16. The learning rate is scheduled from 0.01 following the reset scheme in Section \ref{sec_rest}. The search process takes about 5.5 hours on one GeForce RTX 3090 GPU. However, since we emulate the hardware bit-shift operations with software implementation, the search time actually can be significantly shortened on the dedicated hardware platforms. We will discuss more about the search efficiency in Section \ref{sec_efficiency}. The best cells searched from CIFAR are shown in Appendix \ref{sec_bestcells}.

\noindent\textbf{Evaluation settings.} 
The evaluation network is composed of 20 cells, including 18 normal cells and 2 reduction cells. We set the initial channel size as 36 and optimize the network via the RAdam optimizer \cite{liu2019variance} with an initial learning rate of 0.01 (cosine annealing to 0) and weight decay of 3$e$-4. Following the setting in DeepShift, the netowrk is trained from scratch with bit-shift weights for 200 epochs. The batch size is set as 128. Cutout and drop-path with a rate of 0.2 are used to prevent overfitting. The training accuracy curves can be found in Appendix \ref{sec_trainacctrace}.

\noindent\textbf{Results analysis.} 
Table \ref{tab_cifar} shows the evaluation results on CIFAR 10/100 datasets. The bit-shift networks searched by \NAME achieve 95.58\% and 76.35\% accuracy on CIFAR10 and CIFAR100, respectively. Compared to conventional manually designed CNNs, \NAME models lead to a significant performance improvement in the bit-shift domain, where the prediction accuracy increases (1.69$\sim$8.07)\% on CIFAR10 and (5.71$\sim$18.09)\% on CIFAR100. 
Moreover, the parameter size of searched networks is also much smaller than most conventional CNNs. Hence, in contrast to directly transferring those CNNs into bit-shift counterparts, \NAME is a more efficient approach to generate high-quality bit-shift networks, with the improved accuracy, reduced parameter size and automatic design process. We also compare \NAME with state-of-the-art NAS techniques searched in the real domain, and the results show that our method can find out architectures more compatible to the bit-shift domain. We will discuss more details in Section \ref{sec_nascom}.

\begin{table}[t]
\centering
\resizebox{\linewidth}{!}{
\begin{tabular}{lccccc}
    \hline
    \hline
    \multirow{2}{*}{\textbf{Architecture}} & \multicolumn{2}{c}{\textbf{Acc. (\%)}} & \textbf{Params} &  \textbf{Multi} &  \textbf{Add}  \\
    \cmidrule(lr){2-3}
        &  \textbf{Top-1}  &   \textbf{Top-5}  &  \textbf{(M)}   &  \textbf{(M)}  &  \textbf{(M)}  \\
    \hline
    ResNet18  &  62.25   &  83.79  &   11.7   & 0  & 987 \\
    ResNet50  &   69.04  & 88.61  &   25.8  &   0  & 2053  \\
    VGG16*  & 0.10  &  0.98  &  138.5  & 0 & 8241 \\
    GoogleNet  &   62.81   & 84.81  &  6.6  &  0  & 752 \\
    MobileNet-v2* & 40.03  &  65.13  &   4.7   & 0   &  206 \\
    ShuffleNet-v2* & 37.32  &  62.26  &   7.4   & 0 & 306 \\
    SqueezeNet1\_0 & 29.08  &  51.96  &  3.8 & 0 & 412 \\
    \hline
    NASNet  & 66.24  & 86.24   &  5.6  &  0  &  317 \\
    DARTS-v2  &  64.98 &  85.18  &  4.7   &  0   & 287 \\
    GDAS  &   65.87   &  85.95  &  5.3  &  0 & 291 \\
    DOTS  & 66.36  & 86.23  &   5.2   &  0  & 302 \\
    \hline
    \NAME (Ours) &  67.17  &  87.38  &   5.1   & 0 & 298 \\
    \hline
    \hline
\end{tabular}}
\caption{Evaluation results on ImageNet. Results of conventional CNNs are obtained with the batch size of 1024 (the same as ours), as the batch size of 256 used in original DeepShift code makes the training too slow. *: The results are the highest accuracy in the training while networks fail to converge.}\label{tab_imagenet}
\vspace{-10pt}
\end{table}

\subsection{Evaluation on ImageNet}
\noindent\textbf{Evaluation settings.} 
Following previous works \cite{liu2018darts,dong2019searching}, we construct the network for ImageNet with the best cells searched from the CIFAR dataset. The evaluation follows the ImageNet-mobile setting, in which the input size is 224$\times$224. The network consists of 14 cells (12 normal cells and 2 reduction cells) with the initial channel size of 46. We train the network in the bit-shift domain for 90 epochs with a batch size of 1024. The RAdam optimizer with an initial learning rate of 0.01 (warming up in the first 5 epochs and cosine annealing to 0) is used. The training accuracy curves can be found in Appendix \ref{sec_trainacctrace}.

\noindent\textbf{Results analysis.} 
Table \ref{tab_imagenet} shows the evaluation results on the ImageNet dataset. It can be found that although some conventional CNNs (e.g., ResNet) still perform well when converted to the bit-shift domain, there are many more state-of-the-art CNNs giving much lower prediction accuracy or even failing to converge, including VGG16, MobileNet-v2 and ShuffleNet-v2, whose final top-1 accuracy drops to 0.09\%, 1.18\% and 9.27\%, respectively.
In contrast, \NAME can converge robustly and achieve 67.17\% top-1 accuracy, which is (4.36$\sim$67.07)\% higher than conventional CNNs except ResNet50. Note that the high accuracy of ResNet50 is obtained at the price of much larger parameter size (5$\times$) and more operations (7$\times$).  
Hence, compared to conventional CNNs, bit-shift networks searched by \NAME perform better with fewer parameters and operations. The comparison with previous NAS techniques also shows that \NAME can generate more compatible architectures for bit-shift networks. 
Given all multiplications in networks are replaced with bit shifts, the number of multi-operations would be 0, which greatly reduces the resource cost and speeds up the model inference. 

\begin{table}[t]
\centering
\resizebox{\linewidth}{!}{
\begin{tabular}{lccccc}
\hline
\hline
  \multirow{2}{*}{\textbf{Architecture}}   & \multirow{2}{*}{\textbf{Domain}}  &  \multicolumn{4}{c}{\textbf{Acc. (\%)}} \\
\cmidrule(lr){3-6}
&  & \textbf{C10}  & \textbf{Diff.} & \textbf{C100} & \textbf{Diff.} \\
\hline
\multirow{2}{*}{ResNet18} & R & 94.45 & - &  72.53 & - \\
                  & BS  & 93.20  & -1.25 & 69.11  & -3.42 \\
\hline
\multirow{2}{*}{ResNet50} & R & 95.12 & - &  74.19 & - \\
                  & BS  & 93.89  & -1.23 & 70.65  & -3.54 \\
\hline
\multirow{2}{*}{DARTS(v2)} &R & 96.48 & - & 78.78 & - \\
                  & BS & 94.80 & -1.68 & 75.17  & -3.61 \\             
\hline
\multirow{2}{*}{DARTS-} & R & 95.61 & - & 76.02  & -  \\
                  & BS  & 93.87 & -1.74  & 70.85  & -5.17 \\
\hline
\multirow{2}{*}{DOTS} & R &  96.55  & -  & 78.87 & - \\
                  & BS  & 95.13 & -1.42 & 75.05 & -3.82 \\  
\hline
\multirow{2}{*}{\NAME} & R & 96.19 & - & 78.26 & - \\
                  & BS & \textbf{95.58} & \textbf{-0.61} & \textbf{76.35} & \textbf{-1.91} \\
\hline
\hline
\end{tabular}}
\caption{Model accuracy of various architectures in the real (R) and bit-shift (BS) domains, and their differences (Diff.). The models are trained over CIFAR10 (C10) and CIFAR100 (C100) datasets}\label{tab_nas}
\vspace{-10pt}
\end{table}

\subsection{Real-valued and Bit-shift Network Comparisons} \label{sec_nascom}
We compare the accuracy of the same network trained in the real and bit-shift domains, aiming to investigate the accuracy drop of conventional CNNs and NAS models caused by the bit-shift quantization. Table \ref{tab_nas} shows the results of some representative networks on the CIFAR datasets. Comparison on ImageNet can be found in Appendix \ref{sec_comimage}. We can observe that \NAME not only achieves the highest accuracy of bit-shift networks, but also leads to the smallest accuracy drop (-0.61\% and -1.91\%) when the network is quantized from the real to bit-shift domains. In comparison, conventional CNNs have lower accuracy in the real domain, and the accuracy drops more significantly during the bit-shift quantization. 

We further compare \NAME with previous NAS techniques. From Table \ref{tab_nas}, \NAME is able to obtain network architectures with better performance in the bit-shift domain, even their accuracy in the real domain is slightly lower. It indicates that transferring existing NAS models directly to the corresponding bit-shift networks normally just achieves sub-optimal solutions. The networks searched by \NAME are more compatible to the bit-shift quantization. 

\subsection{Ablation Study} \label{sec_abst}

\noindent\textbf{Impact of the shift-oriented search space.}
The superiority of \NAME in the bit-shift domain actually has indicated the effectiveness of the shift-oriented search space, which avoids converging to sub-optimal solutions for searching bit-shift network architectures. To further validate the importance of this new search space, we replace the search space with the classical real-valued one in \NAME, and then check the performance of the searched results. Four experiments are run individually with random seeds, where the searched architectures achieve average accuracy of 94.97\% on CIFAR10 and 75.03\% on CIFAR100. It drops 0.63\% and 1.32\% from that with the shift-oriented search space. Besides, as a by-product, the shift-oriented search space significantly reduces the resource cost in the search process, as it replaces dense multiplications with much cheaper bit shifts. Hence, \NAME can generate better bit-shift networks automatically with much less resource budget. 

\begin{figure}[t]
  \centering
  \begin{subfigure}[b]{0.48\linewidth}
    \centering
    \includegraphics[width=\linewidth]{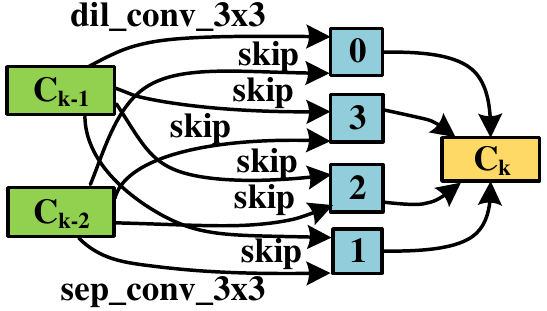}
    \caption{DARTS cell}
    \label{fig_darts}
  \end{subfigure}%
  \hfill
  \begin{subfigure}[b]{0.48\linewidth}
    \centering
    \includegraphics[width=\linewidth]{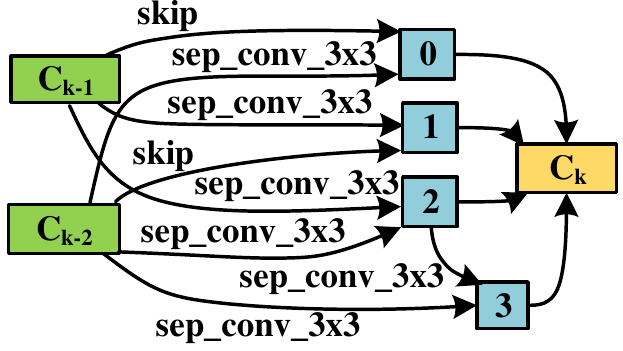}
    \caption{ShiftNAS cell}
    \label{fig_shiftnas}
  \end{subfigure}
  \vspace{-10pt}
  \caption{Normal cell structures searched from the bit-shift domain}
  \vspace{-10pt}
  \label{fig_cells}
\end{figure} 

\noindent\textbf{Impact of the topology-related search strategy.}
We take DARTS as the baseline strategy to derive cell structures from the shift-oriented search space. The result is shown in Figure \ref{fig_darts}. It can be seen that the searched cell is dominated by the skip connections and only achieves 69.58\% accuracy on CIFAR100. This is because the drawback of the traditional gradient-based search strategy is amplified in the bit-shift domain. By integrating our topology-related search strategy, this drawback can be effectively mitigated and the searched result is shown in Figure \ref{fig_shiftnas}. Since the edge connections are further inspected, the topology-related search strategy can generate more stable architectures and achieve 76.21\% accuracy, which is 6.63\% improvement over DARTS.

\begin{table}[t]
\centering
\resizebox{0.8\linewidth}{!}{
\begin{tabular}{ccccccc}
\hline
\hline
 \multirow{2}{*}{\textbf{ID}} & \multicolumn{2}{c}{\textbf{Scheme}}  &  \multicolumn{4}{c}{\textbf{Acc. (\%)}} \\
\cmidrule(lr){2-3}\cmidrule(lr){4-7}
& \textbf{L2R} & \textbf{LRR} & \textbf{C10}  & \textbf{Diff.} & \textbf{C100} & \textbf{Diff.} \\
\hline
1 & \Checkmark & \Checkmark & 95.58 & - &  76.35 & - \\
2 & \Checkmark  & \XSolidBrush  & 95.17  & -0.41 & 73.86  & -2.49 \\
3 &\XSolidBrush & \Checkmark & 95.43 & -0.15 &  74.93 & -1.42 \\
4 &  \XSolidBrush  & \XSolidBrush  & 94.91  & -0.67 &  73.04 & -3.31 \\
\hline
\hline
\end{tabular}}
\caption{Model accuracy for different scheme combinations. Diff. is the accuracy difference from the one with L2R and LRR enabled.}\label{tab_scheme}
\vspace{-10pt}
\end{table}

\begin{figure}[t]
    \centering
    \includegraphics[width=0.85\linewidth]{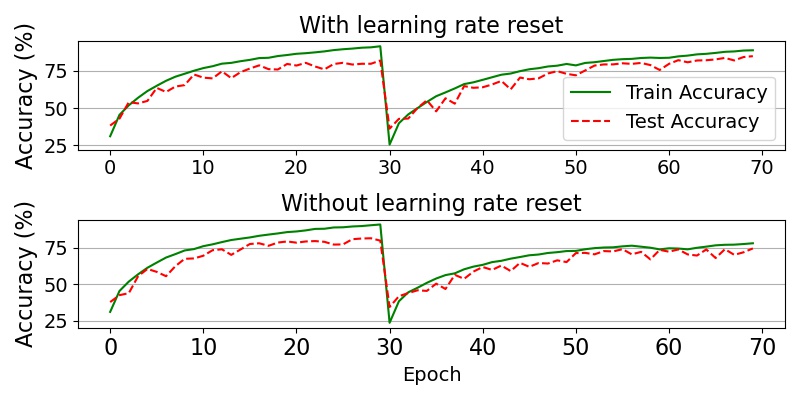}
    \vspace{-10pt}
    \caption{Accuracy trace in the search process on CIFAR10}
    \label{fig_lrrenew}
    \vspace{-15pt}
\end{figure}

\noindent\textbf{Impact of regularization and stabilization.}
To evaluate the effectiveness of our modified \emph{L2 regularization (L2R)} and \emph{learning rate reset (LRR)} schemes, we compare the performance of networks searched with various scheme combinations (Table \ref{tab_scheme}). We find that while both schemes increase the accuracy of the searched architecture, LRR contributes more than L2R. Figure \ref{fig_lrrenew} shows the accuracy curves of the search process on CIFAR10 with or without LRR. It clearly shows that LRR scheme significantly improves the model accuracy from 74.58\% to 84.68\%, which makes it more possible to search for better bit-shift networks. Note that at the start of topology search (the 30\textit{th} epoch), the model gets pruned and retrained, so the accuracy has a sharp drop. 

\subsection{Efficiency Analysis}\label{sec_efficiency}
Given that modern computer architectures use the binary format to store and calculate data, bitwise operations like bit shift and addition are the atomic units for performing complex computations, including the multiplication. According to \cite{cpucycle}, the floating-point multiplication takes at least 5$\times$ of clock cycles than the bit shift.
Besides, compared to the hardware implementation of bit shift on the circuit, the multiplier takes at least 9.7$\times$ of average power, 1.45$\times$ of area and 4.32$\times$ of transistors \cite{asati2009comparative}. Hence, by replacing floating-point weights with bit shift and sign flip operations, the efficiency of architecture search can be significantly improved over previous NAS techniques that search in the real domain. While our software emulation of \NAME just takes 5.5 hours, where the bit shift is simulated by multiplying the power of 2, the actual search cost on the dedicated hardware platforms (e.g., FPGA accelerators) would be largely decreased. We deem that accelerating the NAS process with bit shift on the FPGA board is a promising research direction. Besides, since the searched architectures are trained as bit-shift networks, it also reduces the resource cost and time of model training and inference. 
\NAME also greatly compresses the storage size of searched networks, as it represents model weights with fewer bits (i.e., 5 bits). This promotes the applications of NAS models on the edge devices, where the memory storage and energy consumption are the main constraints.

\section{Conclusion and Future Work} \label{sec_con}
In this paper, we propose to automatically generate advanced bit-shift networks with a dedicated NAS method \NAME. We overcome the challenges of applying existing NAS techniques in the bit-shift domain with three innovations: shift-oriented search space, topology-related search strategy and search regularization and stabilization.
Experimental results show that \NAME can search for architectures with higher compatibility for bit-shift operations, and better performance than state-of-the-art CNNs and NAS models. 

While replacing model multiplications with bit shifts can efficiently reduce the running cost, it is essentially a coarse-grained representation of model weights, which naturally results in the non-trivial drop of prediction accuracy. To address this, we can further introduce additions into the search space of \NAME, which are also efficient substitutes of multiplications \cite{chen2020addernet} and more importantly, can achieve finer-grained weight manipulation \cite{you2020shiftaddnet}. Since current CUDA kernels lack optimization of intensive additions, we leave it as the future work.

\nocite{langley00}

\bibliography{refer}
\bibliographystyle{icml2021}

\newpage
\appendix

\section{Architecture Search Details}
For the operation search, the official CIFAR training dataset is divided into two halves: training set $\mathbb{D}_T$ and validation set $\mathbb{D}_V$, which are used to optimize network weights $w$ and operation weights $\alpha$, respectively. The topology search directly uses the whole official training set to optimize the topology weight $\beta$ with one-level optimization, where the initial temperature $T_0$ is set as 10 and decay to 0.02. We adopt Rectified Adam (RAdam) optimizer with initial learning rate of 0.01 and weight decay of 3e-4 to optimize model weight $w$ and Adam optimizer with initial learning rate of 3e-4 and weight decay of 1e-3 to optimize operation weight $\alpha$ and topology weight $\beta$. The learning rate is scheduled with cosine scheduler following our proposed learning rate reset scheme. The search process consists of 70 epochs with the batch size of 128, including 30 epochs for operation search and 40 epochs for topology search.  

\section{Architecture Evaluation Details}
\noindent\textbf{Training on CIFAR.}
We train the evaluation network for 200 epochs with the batch size of 128. The network is optimized by RAdam optimizer with initial learning rate of 0.01 and weight decay of 3e-4. The learning rate is scheduled by a cosine annealing scheduler to 0. Cutout and drop-path with a rate of 0.2 are used for preventing overfitting. 

\noindent\textbf{Training on ImageNet.}
The network is trained by 90 epochs with the batch size of 1024. The RAdam optimizer is adopted, whose initial learning rate is set as 0.01 and weight decay is set as 3e-4. The learning rate is cosine annealed to 0. Label smoothing and an auxiliary loss tower is used to enhance model training. 

\begin{table}[h]
\centering
\resizebox{\linewidth}{!}{
\begin{tabular}{|c|c|c|c|}
\hline
\textbf{Dataset}  &  \textbf{Cell}  & \textbf{Node} & \textbf{Genotype} \\ \hline
\multirow{8}{*}{CIFAR10} & \multirow{4}{*}{\makecell[c]{Normal \\ Cell}} & 1 & ('skip\_connect', 0), ('skip\_connect', 1) \\ \cline{3-4} 
  &   & 2 & ('sep\_conv\_3x3', 0), ('sep\_conv\_3x3', 1) \\ \cline{3-4} 
  &   & 3 & ('sep\_conv\_3x3', 0), ('sep\_conv\_3x3', 1) \\ \cline{3-4} 
  &   & 4 & ('sep\_conv\_3x3', 0), ('dil\_conv\_5x5', 4) \\ \cline{2-4} 
  & \multirow{4}{*}{\makecell[c]{Reduction \\ Cell}} & 1 & ('skip\_connect', 0), ('skip\_connect', 1)  \\ \cline{3-4} 
  &   & 2 & ('sep\_conv\_3x3', 0), ('max\_pool\_3x3', 1) \\ \cline{3-4} 
  &   & 3 & ('sep\_conv\_3x3', 0), ('sep\_conv\_5x5', 1) \\ \cline{3-4} 
  &   & 4 & ('skip\_connect', 0), ('dil\_conv\_5x5', 2) \\ \hline
\end{tabular}}
\caption{Genotype of Best Archtiecture on CIFAR10}\label{tab_c10best}
\vspace{-10pt}
\end{table}

\begin{table}[h]
\centering
\resizebox{\linewidth}{!}{
\begin{tabular}{|c|c|c|c|}
\hline
\textbf{Dataset}  &  \textbf{Cell}  & \textbf{Node} & \textbf{Genotype} \\ \hline
\multirow{8}{*}{CIFAR100} & \multirow{4}{*}{\makecell[c]{Normal \\ Cell}} & 1 & ('sep\_conv\_3x3', 0), ('skip\_connect', 1) \\ \cline{3-4} 
  &   & 2 & ('skip\_connect', 0), ('sep\_conv\_3x3', 1) \\ \cline{3-4} 
  &   & 3 & ('sep\_conv\_3x3', 0), ('sep\_conv\_3x3', 1) \\ \cline{3-4} 
  &   & 4 & ('sep\_conv\_3x3', 0), ('sep\_conv\_5x5', 4) \\ \cline{2-4} 
  & \multirow{4}{*}{\makecell[c]{Reduction \\ Cell}} & 1 & ('max\_pool\_3x3', 0), ('skip\_connect', 1)  \\ \cline{3-4} 
  &   & 2 & ('sep\_conv\_5x5', 0), ('sep\_conv\_5x5', 1) \\ \cline{3-4} 
  &   & 3 & ('max\_pool\_3x3', 0), ('dil\_conv\_5x5', 3) \\ \cline{3-4} 
  &   & 4 & ('sep\_conv\_5x5', 0), ('sep\_conv\_3x3', 3) \\ \hline
\end{tabular}}
\caption{Genotype of Best Archtiecture on CIFAR100}\label{tab_c100best}
\vspace{-10pt}
\end{table}

\section{Best Searched Cell Structures}\label{sec_bestcells}
Table \ref{tab_c10best} and \ref{tab_c100best} show the best searched architectures for CIFAR10 and CIFAR100. The evaluation on ImageNet adopts cells searched from CIFAR10 (Table \ref{tab_c10best}). 

\section{Training Results}\label{sec_trainacctrace}
\begin{figure}[h]
\centering
\subfloat[CIFAR10]{
  \includegraphics[clip,width=0.9\columnwidth]{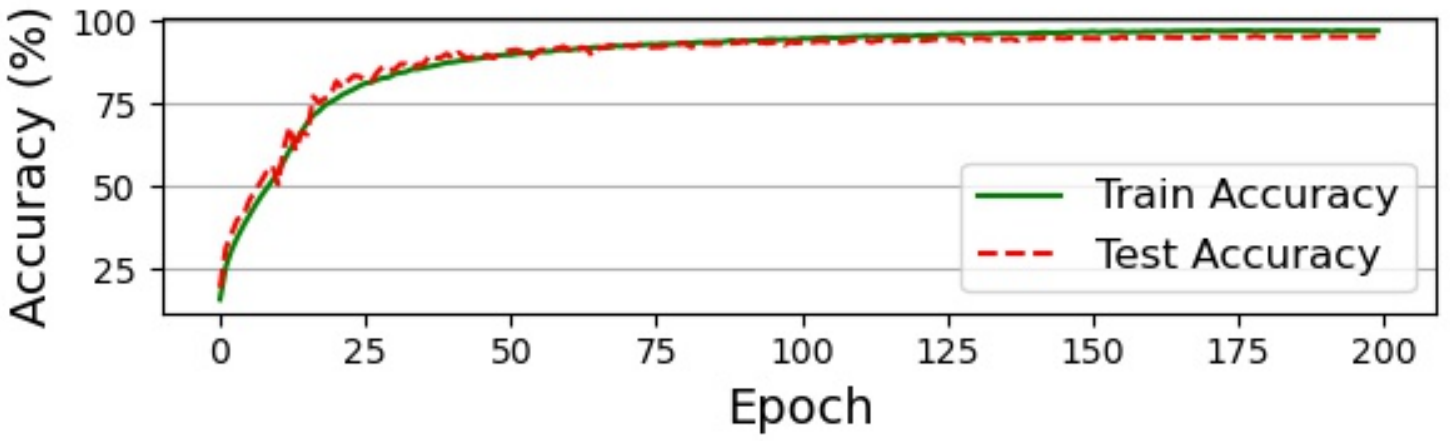}
}\\
\subfloat[CIFAR100]{
  \includegraphics[clip,width=0.9\columnwidth]{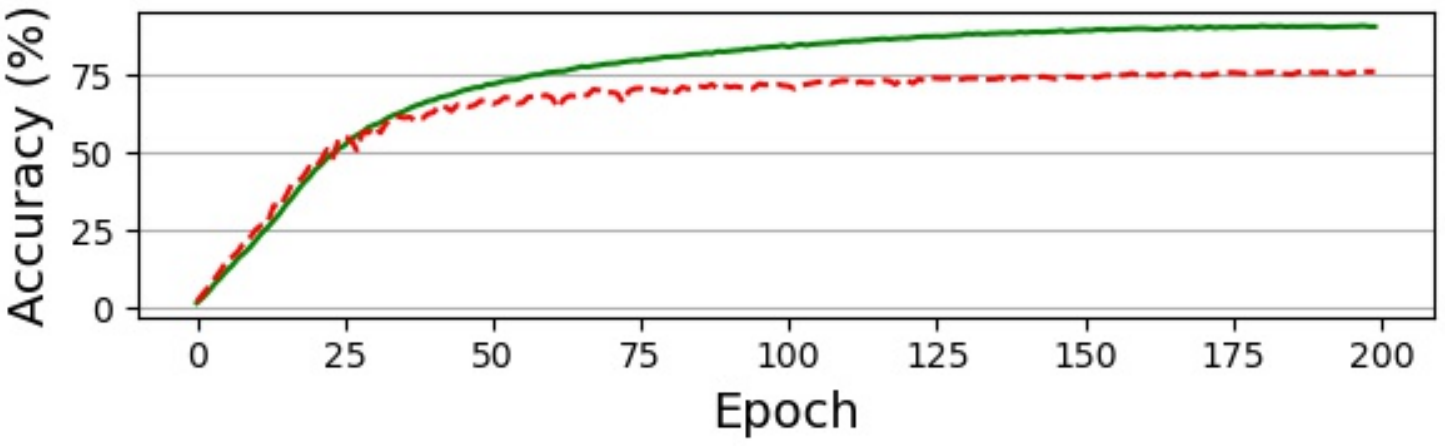}
}
\caption{Training result on CIFAR}\label{fig_cifartrain}
\end{figure}

\begin{figure}[h]
\centering
\subfloat[Batch size = 1024]{
  \includegraphics[clip,width=0.9\columnwidth]{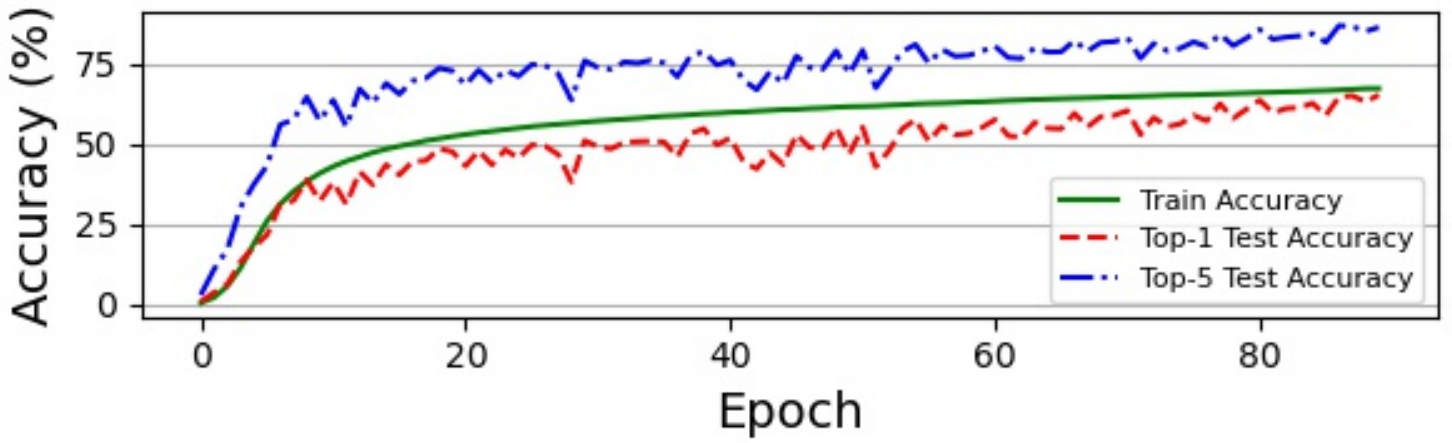}
}\\
\subfloat[Batch size = 256]{
  \includegraphics[clip,width=0.9\columnwidth]{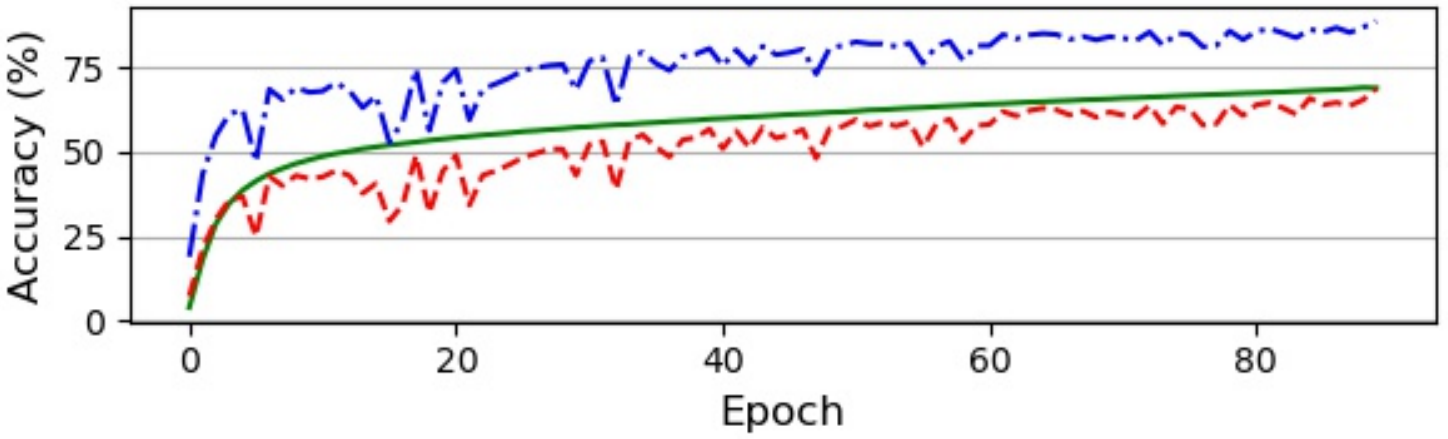}
}
\caption{Training result on ImageNet}\label{fig_imagenettrain}
\vspace{-10pt}
\end{figure}

Figure \ref{fig_cifartrain} shows the accuracy traces of training on CIFAR10 and CIFAR100. Figure \ref{fig_imagenettrain} shows the accuracy traces of training on ImageNet, where (a) takes batch size of 1024 and (b) takes 256. It can be seen that training with batch size of 256 converges earlier and is also more stable, where the final top-1 accuracy is slightly higher (68.67\% vs. 67.17\%).

\section{Comparison with Real-valued Counterparts on ImageNet} \label{sec_comimage}
Due to the limitation of resource and time, we just select each a model from conventional CNNs (i.e., ResNet18) and previous NAS methods (i.e., DOTS) to compare the accuracy drop from the real-valued counterparts on the ImageNet with our proposed \NAME. Table \ref{tab_nasimagenet} shows the results. It can be found that the architecture searched by \NAME achieves the highest accuracy as a bit-shift network, and also has the lowest accuracy drop from the counterpart training in the real domain. Compared to other conventional CNNs and even most state-of-the-art NAS models, ResNet have more robust performance even training with bit-shift weights. However, it is still worse than our proposed \NAME, and more importantly, ResNets are much more heavy than NAS searched models. 

\begin{table}[h]
\centering
\resizebox{\linewidth}{!}{
\begin{tabular}{lccccc}
\hline
\hline
  \multirow{2}{*}{\textbf{Architecture}}   & \multirow{2}{*}{\textbf{Domain}}  &  \multicolumn{4}{c}{\textbf{Acc. (\%) on ImageNet}} \\
\cmidrule(lr){3-6}
&  & \textbf{Top-1}  & \textbf{Diff.} & \textbf{Top-5} & \textbf{Diff.} \\
\hline
\multirow{2}{*}{ResNet18} & R & 68.14 & - &  88.67 & - \\
                  & BS & 62.25  & -5.89 & 83.79  & -4.88 \\
\hline
\multirow{2}{*}{DOTS} & R & 72.75 & - &  90.96 & - \\
                  & BS  & 66.36  & -6.39 & 86.23  & -4.73 \\
\hline
\multirow{2}{*}{ShiftNAS} & R & 72.18 & - & 90.61 & - \\
                  & BS & 67.17 & -5.01 & 87.38  & -3.23 \\             
\hline
\hline
\end{tabular}}
\caption{Model accuracy on ImageNet of various architectures in the real (R) and bit-shift (BS) domains, and their differences (Diff.).}\label{tab_nasimagenet}
\vspace{-10pt}
\end{table}

\end{document}